\documentclass{article} 
\usepackage{iclr2024_conference,times}

\usepackage{amsmath,amsfonts,bm}

\def\eqref#1{equation~\ref{#1}}

\def\1{\bm{1}}

\DeclareMathAlphabet{\mathsfit}{\encodingdefault}{\sfdefault}{m}{sl}
\SetMathAlphabet{\mathsfit}{bold}{\encodingdefault}{\sfdefault}{bx}{n}

\usepackage[utf8]{inputenc} 
\usepackage[T1]{fontenc}    
\usepackage{hyperref}       
\usepackage{url}            
\usepackage{booktabs}       
\usepackage{amsfonts}       
\usepackage{nicefrac}       
\usepackage{microtype}      
\usepackage{xcolor}         

\usepackage{graphicx}       
\usepackage{caption}
\usepackage{algorithm}
\usepackage{algorithmicx}
\usepackage{algpseudocode}
\usepackage{amsmath}
\usepackage{amsthm}
\usepackage{wrapfig}
\usepackage{booktabs}
\usepackage{amssymb}
\usepackage{multirow}
\usepackage{multicol}
\usepackage{enumitem}
\usepackage{caption}
\usepackage{subfigure}
\usepackage{makecell}
\usepackage{pifont}

\newcommand{\ieno}{\textit{i}.\textit{e}.}

\newcommand{\egno}{\textit{e}.\textit{g}.} 
\newcommand{\etc}{\textit{etc}.}
\newcommand{\etcno}{\textit{etc}} 

\newcommand{\ours}{RUIG}

\title{Reinforced UI Instruction Grounding: Towards a Generic UI Task Automation API}

\iclrfinalcopy

\author{%
  Zhizheng Zhang\quad\; ~
  Wenxuan Xie\quad\; ~
  Xiaoyi Zhang\quad\; ~
  Yan Lu\quad\; ~\\
  Microsoft Research Asia \\ 
  \texttt{\{zhizzhang,\;wenxie,\;xiaoyizhang,\;yanlu\}@microsoft.com}
}

\begin{document}

\maketitle

\begin{abstract}
Recent popularity of Large Language Models (LLMs) has opened countless possibilities in automating numerous AI tasks by connecting LLMs to various domain-specific models or APIs, where LLMs serve as dispatchers while domain-specific models or APIs are action executors. Despite the vast numbers of domain-specific models/APIs, they still struggle to comprehensively cover super diverse automation demands in the interaction between human and User Interfaces (UIs). In this work, we build a multimodal model to ground natural language instructions in given UI screenshots as a generic UI task automation executor. This metadata-free grounding model, consisting of a visual encoder and a language decoder, is first pretrained on well studied document understanding tasks and then learns to decode spatial information from UI screenshots in a promptable way. To facilitate the exploitation of image-to-text pretrained knowledge, we follow the \textit{pixel-to-sequence} paradigm to predict geometric coordinates in a sequence of tokens using a language decoder. We further propose an innovative Reinforcement Learning (RL) based algorithm to supervise the tokens in such sequence jointly with visually semantic metrics, which effectively strengthens the spatial decoding capability of the \textit{pixel-to-sequence} paradigm. Extensive experiments demonstrate our proposed reinforced UI instruction grounding model outperforms the state-of-the-art methods by a clear margin and shows the potential as a generic UI task automation API.
\end{abstract}

\section{Introduction}

Interacting with User Interfaces (UIs) pervades most people's daily work and life. These interaction activities are associated with diverse purposes from numerous users, imposing a wealth of achieving UI task automation for improving the interaction efficiency and experiences. This is in fact especially urgent for disabilities and is in line with the spirit of AI for Good.

The success of advanced Large Language Models (LLMs) \citep{ChatGPT,GPT4,touvron2023llama,chung2022scaling,zhang2022opt} has been opening countless possibilities for task automation by taking advantage of generic procedural knowledge in LLMs. Recently, there is a surge of research works \citep{ChatGPTplugins,AutoGPT,AgentGPT,vemprala2023chatgpt,yang2023mm,shen2023hugginggpt,liang2023taskmatrix,wu2023visual} dedicated to automating AI tasks with the collaboration between LLMs and various domain-specific models or APIs. In these paradigms, LLMs function as planners to parse task goals into a sequence of executable commands, where the task goals are high-level instructions from humans while the executable commands are low-level instructions generated by LLMs and fed into executors for execution in practice. The executors here could be plugins \citep{ChatGPTplugins}, curated tools \citep{AutoGPT,AgentGPT}, AI models \citep{shen2023hugginggpt,wu2023visual} or APIs \citep{liang2023taskmatrix,yang2023mm,yang2023mm}. However, to the best of our knowledge, none of the existing models are competent enough to cover rich requirements for the executors in UI task automation since this field involves a wide range of application scenarios across diverse user intentions and software platforms.

In the field of UI task automation, there are previous efforts \citep{gur2018learning,liu2018reinforcement,humphreys2022data,iki2022berts,li2020widget,kim2023language} dedicated to learning to control computers on a suite of website browsing tasks in simulated environments, \egno, MiniWoB \citep{shi2017world}, MiniWoB++ \citep{liu2018reinforcement}, \etcno. However, the UIs in the real world have more diverse and complicated layouts with more UI elements compared to those in simulated environments. To target the challenges in the real world, recent advanced works \citep{li2020mapping,he2021actionbert,bai2021uibert,burns2022dataset,li2022spotlight} learn to ground the target element associated with the given instructions. These methods requires the metadata \citep{li2020mapping,burns2022dataset} (\egno, view hierarchies) or additional information \citep{he2021actionbert,bai2021uibert,li2022spotlight} (\egno, the bounding boxes of UI elements) as the inputs for grounding the target UI element, which limits their practical use. This is because the metadata and the additional information are not always available, and the quality of metadata provided by third-party developers is hard to guaranteed.
In this paper, we propose a powerful generic UI instruction grounding model that only requires the text-represented instructions and screenshots as its inputs, obviating the need for metadata or any other additional information. 

UI screenshots contain rich and dense visual and textual information. UI instruction grounding aims to localize the target element at each step for automatically completing clicking or typing operations in line with human instructions. Its core challenge lies in learning not only precise but also dense correlations between textual information in instructions and visual information in screenshots. Besides, the relative relations between densely arranged UI elements also need to be captured. Admittedly, this task is challenging, the core knowledge it requires has been learned in part by full-fledged image-to-text models, such as document understanding \citep{kim2022ocr,xu2020layoutlm} models. \textit{Could we unleash inherent capabilities of these full-fledged models for building a high-performance instruction grounding model?}   

An intuitive way is to treat aforementioned models as the pre-trained models and perform fine-tuning on our targeted task. These models take images as inputs while generating the outputs in linguistic form, constraining us to model the outputs of our targeted instruction grounding in linguistic form as well. Recent novel \textit{pixel-to-sequence} based works \citep{chen2021pix2seq,chen2022unified,yang2022unitab} inspire us to localize the target UI element by predict its bounding box in linguistic form. However, unfortunately, it is not easy as expect to attain favorable performance on our targeted task straightforwardly. This is because language decoders generate a sequence autoregressively where each token is supervised independently rather than adopting a training loss jointly for a set of tokens corresponding to bounding box coordinates. It in fact exposes a limitation of the \textit{pixel-to-sequence} paradigm: the model has no awareness of the combinational semantics for a set of tokens. In our targeted problem, such combinational semantics refers to the visual geometric properties of the target bounding box. In this paper, we propose a policy gradients \citep{sutton2018reinforcement} based approach to break through this limitation and enhance the spatial decoding capability of \textit{pixel-to-sequence} paradigm by supervising a set of tokens in a sequence jointly. It enables us to train a powerful UI instruction grounding model towards a generic UI task automation API. We name it Reinforced UI instruction grounding (\ours) model.

We summarize the contributions of this work as follows:
\begin{itemize}[leftmargin=*,noitemsep,nolistsep]
    \item We construct a powerful UI instruction grounding model, dubbed \ours, that only requires text instructions and screenshots as the inputs, circumventing the need for the metadata of UIs or other additional information. It could serve as a generic UI task automation execution API.
    \item We propose a policy gradients based approach to endow the training of \textit{pixel-to-sequence} paradigm with the awareness of the combinational semantics of its decoded token sequence. It enables our proposed \ours~model to be capable of taking into account the visual geometric properties of the positions of target UI elements when learning to decode them in linguistic form.
    \item We conduct extensive experiments to demonstrate the effectiveness of our proposed \ours~and show it can outperform the state-of-the-arts, including the metadata-involved ones, by a clear margin.  
\end{itemize}

\section{Related Works}
\subsection{Instruction Grounding}

In the era of LLMs, LLMs have exhibited impressive capabilities of planning high-level instructions from human into executable low-level (step-wise) instructions \citep{AutoGPT,AgentGPT,vemprala2023chatgpt,shen2023hugginggpt,liang2023taskmatrix,kim2023language}, in urgent need of a powerful instruction grounding model as a expert executor for UI task automation. Instruction grounding is at the core of automated action execution in UI tasks by localizing the target UI elements upon the given step-wise instructions. Once given the locations of target UI elements, practical mouse or keyboard operations can be easily achieved by open-sourced tools, \egno, PyAutoGUI\footnote{\url{https://pyautogui.readthedocs.io/en/latest/}}. Many previous efforts \citep{gur2018learning,liu2018reinforcement,humphreys2022data,iki2022berts,li2020widget,kim2023language} are made for learning to automatically control computers on website browering tasks in simulated environments, \egno, MiniWoB \citep{shi2017world}, MiniWoB++ \citep{liu2018reinforcement}, \etcno. Recent research works \citep{li2020mapping,he2021actionbert,bai2021uibert,burns2022dataset,li2022spotlight} strive for a further step in this field by investigating this topic on real-world mobile data. These methods require the metadata \citep{li2020mapping,burns2022dataset} (\egno, view hierarchies) or additional information \citep{he2021actionbert,bai2021uibert,li2022spotlight} (\egno, the bounding boxes of UI elements) as model inputs. Besides this availability issue, their performance heavily rely on the quality of these information. Towards a generic solution, we propose a UI instruction grounding model which only takes natural language instructions and vision screenshots as inputs, obviating the needs for any metadata or additional information.

\subsection{Pixel-to-sequence Paradigm}

Recently, a big convergence on Vision-Language (VL) tasks \citep{chen2021pix2seq,chen2022unified,yang2022unitab,cho2021unifying,gupta2022towards,jang2022unifying} is gradually formed by unifying multiple VL tasks into a single model against the proliferation of various model designs. Among them, \textit{pixel-to-sequence} \citep{chen2021pix2seq,chen2022unified,yang2022unitab} is a newly devised paradigm of translating vision inputs into discrete tokens, \ieno, decoding bounding boxes, key points, captions, \etc, in linguistic form. We apply the spirit of \textit{pixel-to-sequence} paradigm to distill a well-trained document understanding model as the pre-trained knowledge for our targeted UI instruction grounding task.
 
\subsection{Reinforcement Learning in CV and NLP}

Reinforcement learning (RL) has been applied to a broad range of research fields, including Computer Vision (CV) \citep{lin2021selectaugment,mathe2016reinforcement,le2022deep,pinto2023tuning} and Natural Language Processing (NLP) \citep{uc2023survey,ramamurthy2022reinforcement,ouyang2022training,ChatGPT}. It plays diverse roles, such as selecting samples for data augmentation \citep{lin2021selectaugment}, designing task-specific algorithms \citep{mathe2016reinforcement}, enhancing fine-tuning performance \citep{pinto2023tuning}, aligning model outputs with human preferences with human feadbacks \citep{ramamurthy2022reinforcement,ouyang2022training,ChatGPT} and more. With a different purpose with these works, in this work, we adopt a policy gradients RL algorithm to endow the \textit{pixel-to-sequence} paradigm with the awareness on the combinational semantic of a set of discrete tokens during its training. It significantly enhances the model performance on our targeted task. We believe this reinforced \textit{pixel-to-sequence} paradigm can be extended more broadly.

\section{Reinforced UI instruction grounding}

\subsection{Preliminary}

UI instruction grounding aims to localize the target UI element in the current UI page based on a given instruction. It can be formulated with a conditional probability $P(\mathbf{e}_t|\mathbf{x},\mathbf{c})$, where $\mathbf{e}_t$, $\mathbf{x}$ and $\mathbf{c}$ denotes the target UI element, the current UI page and the text-represented instruction, respectively. In previous works, the UI page $\mathbf{x}$ is described by textual meta data \citep{li2020mapping,burns2022dataset}, element-wise visual patches from screenshots \citep{he2021actionbert,bai2021uibert}, the UI screenshot and a region of interest on the screen \citep{li2022spotlight}. They commonly model $p(\mathbf{e})$ as the probability that $\mathbf{e}$ is the target element conforming to the given instruction where one with the largest probability is the localization result. The bounding boxes of all UI elements are required as priors for these methods when learning $P(\mathbf{e}_t|\mathbf{x},\mathbf{c})$, limiting their generic using in practice. In this work, we introduce a powerful model (named \ours) for this task which directly predicts the bounding box of the target UI element from the screenshot of the current UI page and the given instruction, obviating the need for metadata and additional information, \egno, bounding boxes of all elements or a region of interest. 

\subsection{Framework Design}

In this section, we introduce the framework of our proposed \ours~model. Overall, \ours~model is an reinforced instantiation of \textit{pixel-to-sequence} paradigm for UI instruction grounding. This reinforced instantiation provides insights from two aspects: 1) It takes advantage of the functionality of \textit{pixel-to-sequence} on unifying the forms of model outputs, allowing to obtain pre-trained knowledge from UI instruction grounding from caption-like models. 2) It enhances the fine-tuning performance of a \textit{pixel-to-sequence} model by injecting the awareness of combinational semantics to its fine-tuning supervisions with policy gradients, which will be detailed in the next section. 

\begin{figure}[t]
	\begin{center}
		\includegraphics[width=\textwidth]{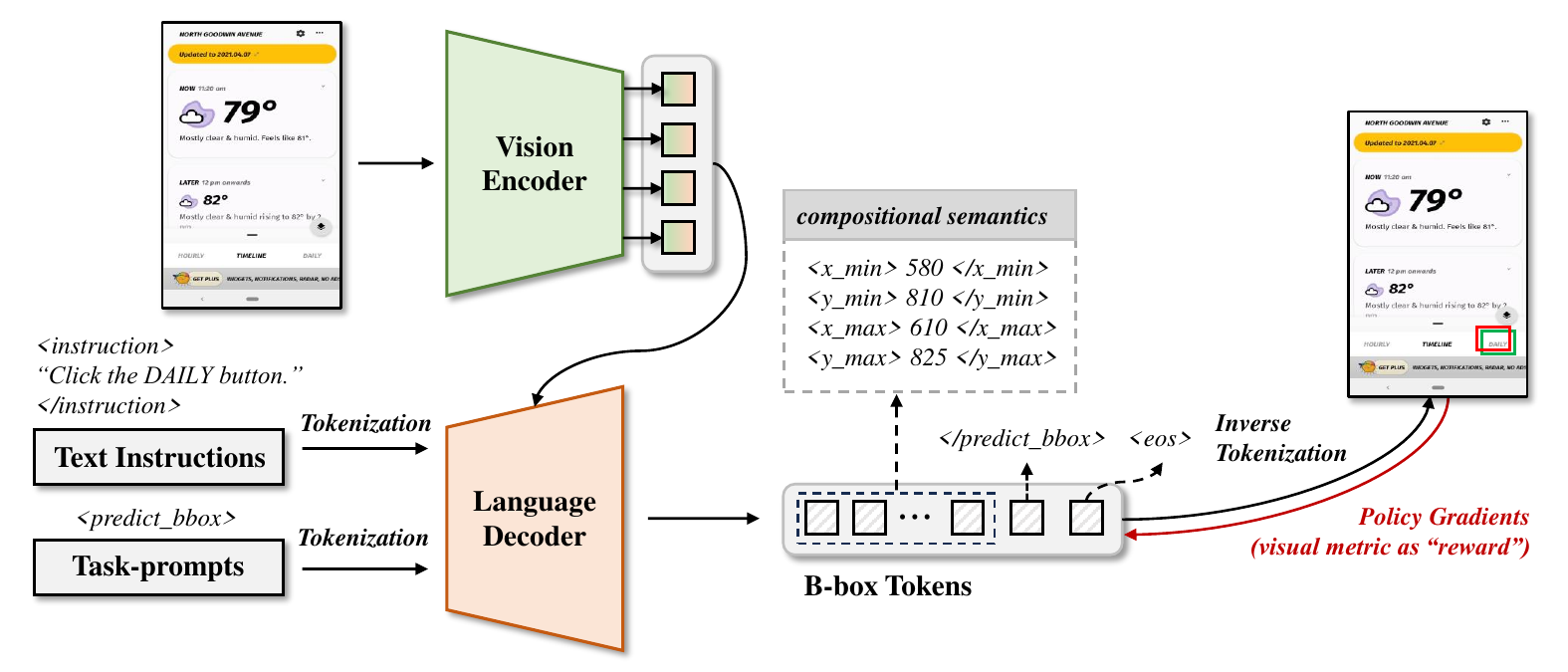} 
	\end{center}
    \vspace{-2mm}
	\caption{The framework of the proposed \ours~model. It consists of a transformer-based vision encoder and a transformer based language decoder, following \textit{pixel-to-sequence} paradigm design. Given a image, it autoregressively decodes the target bounding box coordinates in linguistic form.}
	\label{fig:framework}
\end{figure}

As illustrated in Figure \ref{fig:framework}, \ours~model consists of a transformer-based vision encoder and a transformer-based language decoder. Given a screenshot $\mathbf{x}\in \mathbb{R}^{H \times W \times C}$, the vision encoder embeds $\mathbf{x}$ as a set of $d$-dimensional tokens $\{\boldsymbol{x}_i | \boldsymbol{x}_i \in \mathbb{R}^d, 1 \leq i \leq {N}_{x}\}$ where $i$ indexes the tokens and $N_{x}$ denotes the number of tokens. The language decoder adopts an off-the-shelf tokenizer to embed the given text instruction $\mathbf{c}$ and a task prompt ``\textit{<predict\_bbox>}'' into another set of $d$-dimensional tokens $\{\boldsymbol{c}_j | \boldsymbol{c}_j \in \mathbb{R}^d, 1 \leq j \leq {N}_{c}\}$ and $\boldsymbol{y}_1 \in \mathbb{R}^d$, respectively. The symbol $j$ indexes $\boldsymbol{c}_j$, and ${N}_{c}$ represents the number of tokens in the instruction token set. Here, the instruction $\mathbf{c}$ has a general format as ``\textit{<instruction> \{content\} </instruction>}'' in which ``\textit{<instruction>}'', ``\textit{\{content\}}'' and ``\textit{</instruction>}'' denote the start, specific content and end of the instruction, respectively, as the example shown in Figure \ref{fig:framework}. The decoder predicts the bounding box coordinates of the target UI element in an autoregressive way, as formulated below:
\begin{equation} \label{eq:1}
    \boldsymbol{y}_n \sim p(\boldsymbol{y}_n | \boldsymbol{x}_{1:N_x}, \boldsymbol{c}_{1:N_c}, \boldsymbol{y}_{1:n-1}),
\end{equation}
where $\boldsymbol{x}_{1:N_x}$ and $\boldsymbol{c}_{1:N_c}$ represent aforementioned vision tokens and textual instruction tokens, respectively. $\boldsymbol{y}_n$ denotes the prediction result for the $n$-th token in the decoded sequence $\{\boldsymbol{y}_n | \boldsymbol{y}_n \in \mathbb{R}^d, 1 \leq n \leq {N}_{y}\}$. The decoded sequence has ${N}_{y}$ tokens in total, including the tokens for task beginning prompt ``\textit{<predict\_bbox>}'', bounding box coordinates of the target UI element, task ending prompt ``\textit{</predict\_bbox>}'' and ``\textit{<eos>}'' in sequence. As shown in Figure \ref{fig:framework}, each bounding box is described as the coordinates of its upper left point and lower right point, \ieno, $[x_{min}, y_{min}, x_{max}, y_{max}]$. Each coordinate is formatted in linguistic form together with its corresponding beginning and ending prompts, \egno, $x_{min}$ is formatted as ``\textit{<x\_min> \{$x_{min}$\} </x\_min>}'' where ``\textit{\{$x_{min}$\}}'' is the value.

\subsection{Pixel-to-sequence Paradigm Meets Policy Gradients}
\label{sec: reinforced_pix2seq}

As formulated in Eq. \ref{eq:1}, our \ours~model follows \textit{pixel-to-sequence} paradigm to decode predicted bounding box coordinates of the target UI element and corresponding prompts as a sequence, and advances it with policy gradients based optimization yielding an improved version. We detail it as follows by providing a unified formulation for \textit{pixel-to-sequence} paradigm, analyzing the limitation of its vanilla version and introducing our improved version in our proposed \ours~model.

\paragraph{A unified formulation for \textit{pixel-to-sequence}.} The training objectives of existing \textit{pixel-to-sequence} methods \citep{chen2021pix2seq,chen2022unified,yang2022unitab} are to maximize the likehood of each expected token based on the conditional and preceding tokens over the decoding sequence, which can be formulated in a unified form as: 
\begin{equation} \label{eq:2}
    maximize\; \sum_{n=2}^{N_y} \mathbf{E}_{\hat{P}}[\log P(\boldsymbol{y}_n | \boldsymbol{x}_{1:N_x}, \boldsymbol{c}_{1:N_c}, \boldsymbol{y}_{1:n-1})],
\end{equation}
where $\mathbf{E}_{\hat{P}}[\cdot]$ is the expectation operator with respect to the distribution $\hat{P}$. Here, $\hat{P}$ is the expected distribution (\ieno, ground-truth distribution) of $P$. $\mathbf{E}_{\hat{P}}[\cdot]$ is commonly implemented by a cross-entropy function between $P$ and $\hat{P}$. $\boldsymbol{x}_{1:N_x}$ and $\boldsymbol{c}_{1:N_c}$ are the vision tokens of the input image and the textual tokens of the input text, respectively. Note that $\boldsymbol{c}_{1:N_c}$ are optional in Eq. \ref{eq:2}, which only exist in multi-modal tasks. 

\paragraph{Limitation of vanilla \textit{pixel-to-sequence}.} The discrete tokens in the decoded sequence $\boldsymbol{y}_{1:N_y}$ have their individual semantics. Each token corresponds to an item of specific linguistic semantics in the token vocabulary. Here, we conceptualize ``combinational semantics'' that refers to the high-level semantics of the combinations of multiple correlated tokens. For example, in our modelling for instruction grounding, the tokens correlated to the values of $(x_{min}, y_{min}, x_{max}, y_{max})$ can describe the location of the target UI element in a joint manner. In \textit{pixel-to-sequence} paradigm, visual characteristics, \egno, the geometric precision of a predicted bounding box, are commonly reflected through such combinational semantics. However, as indicated by Eq. \ref{eq:2}, vanilla \textit{pixel-to-sequence} models maximize the likehood of the expected tokens in a token-wise way, lacking the awareness of combinational semantics during model training. 

\paragraph{Reinforced \textit{pixel-to-sequence} model.}

In fact, it is not easy as expect to inject aforementioned combinational semantics into the optimization of a \textit{pixel-to-sequence} based model, \egno, directly maxmizing the IoU metric \citep{zhou2019iou}, as the decoding is autoregressive and the inverse tokenization is not differentiable. In our proposed \ours~model, we model combinational semantics as a reward signal and maximize this reward by adopting policy gradients \citep{sutton2018reinforcement}, \ieno, performing optimization with the gradients of rewards with respect to network parameters. Mathematically, its training objective can be formulated as:
\begin{equation} \label{eq:3}
    maximize\; \sum_{n=2}^{N_y}\nabla_\theta \mathbf{E}_{p}[R(\mathcal{D}_{\boldsymbol{y}_n})] = \sum_{n=2}^{N_y} \mathbf{E}_{p}[R(\mathcal{D}_{\boldsymbol{y}_n})\cdot \log P(\boldsymbol{y}_n | \boldsymbol{x}_{1:N_x}, \boldsymbol{c}_{1:N_c}, \boldsymbol{y}_{1:n-1}; \theta)],
\end{equation}
where $\mathcal{D}_{\boldsymbol{y}_n}$ denotes a set of tokens that share the same combinational semantics with $\boldsymbol{y}_n$, and $R(\mathcal{D}_{\boldsymbol{y}_n})$ refers to the reward for the token $\boldsymbol{y}_n$ calculated over $\mathcal{D}_{\boldsymbol{y}_n}$. The symbol $\theta$ denotes network parameters.

In our proposed \ours~model, we adopt a policy gradients based algorithm for directly maxmizing the IoU metric between the predicted bounding box and its ground-truth. It offers our model with the awareness of the combinational semantics on bounding boxes during training, yielding better alignment between the training of this \textit{pixel-to-sequence} model and the task goal. In our model, the decoded sequence 
includes the tokens for task prompts, coordinate prompts, coordinate values and a end mark of decoding. The reward $R(\mathcal{D}_{\boldsymbol{y}_n})$ is modeled as a vanilla IoU metric for the tokens corresponding to coordinate values (\ieno, $\mathcal{D}_{\boldsymbol{y}_n}$) while being set to zero for other tokens. All tokens in $\mathcal{D}_{\boldsymbol{y}_n}$ share the same reward value. We estimation the expectation value in Eq. \ref{eq:3} via Monte Carlo sampling as common practices in RL field. The \ours~model is finally trained with the objectives in Eq. \ref{eq:2} and Eq. \ref{eq:3} together. We evaluate the effectiveness of our proposed method on UI UI instruction grounding with extensive experiments in the next section, and hope it can inspire broader extensions to more tasks in the future.

\section{Experiments}

\subsection{Experiment Setup}

\paragraph{Datasets.} In this paper, we conduct experiments on both mobile and desktop data. For the experiments with mobile data, we employ a newest benchmark proposed in \citep{burns2022dataset} and follow its corresponding configurations. This benchmark work introduces a new dataset named MoTIF, and proposes a configuration that combining a existing dataset RicoSCA \citep{li2020mapping} and partial MoTIF for training while adopting a sub-set of MoTIF for testing. With this configuration, two experiment settings that have different training-testing splits. In the APP seen setting, the APPs that appear in the test split are all included into those in the train split. In the APP unseen setting, there is no APP overlap between the train and test splits. As for the experiments with desktop data, we collect about 37K UI images from Common Crawl\footnote{\url{https://commoncrawl.org/}}, an open repository of web crawl data. We follow the practices in the open repository\footnote{\url{https://github.com/aburns4/MoTIF}} of \citep{burns2022dataset} to generate 0.5M image-instruction pairs and their corresponding ground-truth labels for instruction grounding task. Similar to the split settings on mobile dataset, we also configure Web seen setting and Web unseen setting on this web crawl dataset for comprehensive evaluation. The data statistics under different settings and the detailed introduction for our web data collection are placed in our supplementary.

\paragraph{Implementation details.} For our proposed \ours~model, we adopt Swin Transformer \citep{liu2021swin} as its vision encoder and employ BART model \citep{lewis2019bart} as its language decoder following \citep{kim2022ocr}. We initialize the entire model weights with those pretrained on a document understanding task, \ieno, captioning all texts in given images from top-left to bottom-right, from \citep{kim2022ocr}. The input resolutions (height $\times$ width) for mobile data and desktop data are $960 \times 640$ and $960 \times 1280$, respectively. The batch size per GPU is set to 3 and 2 for the training on mobile data and desktop data, respectively. We use Adam optimizer to train each model for 20 epochs on 8 NVIDIA V100 GPUs. The initial learning rate is set to $1\!\times\!10^{-4}$ and a cosine learning rate annealing schedule is adopted. The weights for training objectives Eq.\ref{eq:2} and Eq. \ref{eq:3} are set to 1 for them both. Unless specifically stated, we perform Monte Carlo sampling 64 times for each expectation term in Eq. \ref{eq:3}. More details are in the supplementary.

\paragraph{Evaluation metrics.} We calculate the task accuracy (abbreviated as ``Acc'') as the proportion of correctly localizing target UI elements by the tested model over all image-instruction pairs in the test splits. Specially, for those models predicting the bounding box of the target boxes, we view the center of the predicted bounding box as the click point and consider a localization process as correct when this predicted point is within the ground-truth bounding box (available in metadata) of the target UI element otherwise incorrect. Besides, we additionally adopt their mIoU scores for evaluating the spatial localization capability of them.

\begin{table}[t]
  \centering
  \caption{Effectiveness evaluation results of our proposed \ours~model. Here, ``Baseline'' refers to the vanilla \textit{pixel-to-sequence} model \citep{chen2021pix2seq} without our proposed policy gradients based optimization. ``w/o'' is short for ``without'', and ``w/o pre-train'' means that we do not utilize the model weights pre-trained on document understanding tasks \citep{burns2022dataset} to initialize the model weights.}
  \resizebox{\textwidth}{!}{
    \begin{tabular}{clcccccccc}
    \toprule
    \specialrule{0em}{1.5pt}{1pt}   
    \multicolumn{2}{c}{\multirow{3}[0]{*}{\textbf{Models}}} & \multicolumn{4}{c}{\textbf{Mobile Data}} & \multicolumn{4}{c}{\textbf{Desktop Data}} \\
    \multicolumn{2}{c}{} & \multicolumn{2}{c}{\textbf{App Seen}} & \multicolumn{2}{c}{\textbf{App Unseen}} & \multicolumn{2}{c}{\textbf{Web Seen}} & \multicolumn{2}{c}{\textbf{Web Unseen}} \\
    \multicolumn{2}{c}{} & mIoU  & Acc (\%) & mIoU  & Acc (\%) & mIoU  & Acc (\%)  & mIoU  & Acc (\%) \\
    \hline
    \specialrule{0em}{1.5pt}{1pt}   
    \multirow{2}[0]{*}{w/o pre-train} & Baseline & 0.46  & 57.78 & 0.31  & 43.53 & 0.37  & 43.39 & 0.35  & 40.50 \\
          & RUIG (Ours) & 0.51  & 66.25 & 0.39  & 58.67 & 0.46  & 52.91 & 0.43  & 50.15 \\
    \hline
    \specialrule{0em}{1.5pt}{1pt}   
    \multirow{2}[0]{*}{with pre-train} & Baseline & 0.52  & 72.23 & 0.42  & 65.03 & 0.45  & 48.69 & 0.41  & 46.46 \\
          & RUIG (Ours) & 0.62  & 81.16 & 0.48  & 73.92 & 0.51  & 61.78 & 0.49  & 59.03 \\
    \bottomrule
    \end{tabular}%
  }
  \label{tab:effectiveness}%
\end{table}%

\subsection{Ablation Study}

\paragraph{Effectiveness of our proposed method.} We evaluate the effectiveness of our proposed method from two aspects: 1) Whether it can break through the aforementioned limitation of \textit{pixel-to-sequence} paradigm \citep{chen2021pix2seq} on our targeted task? 2) Is it an effective scheme in exploiting pre-trained knowledge from full-fledged document understanding models for constructing high-performance metadata-free UI instruction grounding models? The related experiment results are reported in Table \ref{tab:effectiveness}. 

In Table \ref{tab:effectiveness}, we observe that our proposed model outperforms the vanilla \textit{pixel-to-sequence} baseline by clear margins over different settings on both mobile and desktop data, either with or without exploiting the model weights pre-trained on document understanding tasks for initialization. Specifically, it attains 8.47\%, 15.14\%, 9.52\% and 9.65\% on \textit{App Seen}, \textit{App Unseen}, \textit{Web Seen}, \textit{Web Unseen} respectively without pre-trained weights, and yields 8.93\%, 8.89\%, 13.09\% and 12.57\% under these settings respectively upon pre-trained weights. These improvements demonstrate the effectiveness of endowing \textit{pixel-to-sequence} paradigm with the awareness of combinational semantics inherently carried by its decoded tokens during the model optimization process. We believe this modification is generally applicable for other tasks, and hope its core idea can inspire more works in the future. Besides, we also observe that the utilization of pre-trained weights bring consistent benefits for both the vanilla \textit{pixel-to-sequence} baseline and our proposed model. This is because our proposed model inherits the core spirit of \textit{pixel-to-sequence} as an reinforced version, and demonstrates the rationality of unleashing full-fledged image-to-text models on our targeted problem.

\paragraph{The granularity of combinational semantics.}

\begin{table}[t]
  \centering
  \caption{Comparison results (Acc, \%) of adopting combinational semantics with different granularities in optimizing our proposed \ours~models. ``PG'' is shot for ``policy gradients''. \textit{Base-CenterPoint} represents the vanilla \textit{pixel-to-sequence} model that directly predicts the coordinates of the center point of the target UI element. \textit{Base-Vertices/B-box} denotes the vanilla \textit{pixel-to-sequence} model that predicts the coordinates of the top-left and bottom-right points of the target UI element. \textit{RUIG-CenterPoint} and \textit{RUIG-Vertices} adopt point-level combinational semantics to the training by calculating the rewards as the Euclidean distance between the predicted point coordinates and its ground-truth coordinates. \textit{RUIG-B-box} adopts the combinational semantics at the bounding box level as recommended.}
  \resizebox{\textwidth}{!}{
    \begin{tabular}{lcccccc}
    \toprule
    \multicolumn{1}{c}{\multirow{2}[0]{*}{\textbf{Models}}} & \multicolumn{1}{c}{\multirow{2}[0]{*}{\textbf{\makecell[c]{PG-based\\Training}}}} & \multicolumn{1}{c}{\multirow{2}[0]{*}{\textbf{\makecell[c]{Granularity}}}} & \multicolumn{2}{c}{\textbf{Mobile Data}} & \multicolumn{2}{c}{\textbf{Desktop Data}} \\
          &       &       & \textbf{App Seen} & \textbf{App Unseen} & \textbf{Web Seen} & \textbf{Web Unseen} \\
    \hline
    \specialrule{0em}{1.5pt}{1pt}   
    Base-CenterPoint & \ding{55} & Token & 74.25 & 66.75 & 49.41 & 48.47 \\
    Base-Vertices/B-box & \ding{55} & Token & 72.23 & 65.03 & 48.69 & 46.46 \\
    RUIG-CenterPoint & \ding{51} & Point & 79.94 & 71.88 & 59.39 & 57.65 \\
    RUIG-Vertices & \ding{51} & Point & 78.92 & 69.18 & 56.85 & 55.49 \\
    RUIG-B-box  & \ding{51} & B-box & 81.16 & 73.92 & 61.78 & 59.03 \\
    \bottomrule
    \end{tabular}%
  }
  \label{tab:granularity}%
\end{table}%

In Sec. \ref{sec: reinforced_pix2seq}, we conceptualize ``combinational semantics'' that refers to the high-level semantics of the combinations of multiple relate tokens. The combinational semantics exit at different granularities. For example, the tokens correlated to $(x, y)$ describe the spatial position of a point while the token correlated to $(x_{min}, y_{min}, x_{max}, y_{max})$ describe the location of a bounding box. In fact, the basic training objective formulated in Eq. \ref{eq:2} consider token-level semantics during the optimization, while our proposed training objective as Eq. \ref{eq:3} considers the semantics of decoded tokens at a higher level than that in Eq. \ref{eq:2}, yielding a more global supervision. Here, we experimentally investigate the impacts of such granularity for optimization.

In Table \ref{tab:granularity}, \textit{RUIG-CenterPoint}, \textit{RUIG-Vertices} and \textit{RUIG-B-box} involve combinational semantics beyond token-level semantics in their training objectives. They are all clearly superior to their corresponding baselines across different settings, demonstrating the effectiveness of injecting combinational semantics into training objectives. Besides, we observe that \textit{Base-Vertices/B-box} is slightly inferior to \textit{Base-CenterPoint}, which in fact exposes the limitation of vanilla \textit{pixel-to-sequence} paradigm in decoding the objectives requiring combinational semantics. \textit{RUIG-B-box} delivers the highest accuracy.
This demonstrates the effectiveness of the supervisions at the most global granularity, and indicates that predicting the bounding box of the target UI element is a reliable modelling for UI element localization. We also note that \textit{RUIG-Vertices} performs the worst. 
This is because the UI elements are manually designed in common so that their boundaries are not easy to be clearly determined thus imposing significant challenges in localizing the vertices without global awareness of its entire region.

\paragraph{Which tokens should be optimized with policy gradients?}

As introduced in Sec. \ref{sec: reinforced_pix2seq}, the reward $R(\mathcal{D}_{\boldsymbol{y}_n})$ in Eq. \ref{eq:3} is modeled as a vanilla IoU metric for the tokens corresponding to coordinate values while being set to zero for other tokens. Here, we compare this proposed practice with that back-propagates the IoU-based rewards to all decoded tokens in the sense that the prompt tokens share the same combinational semantics with the coordinate value tokens. As shown in Table \ref{eq:3}, \textit{RUIG (all tokens)} can still achieve improvements relative to the baseline model, but is inferior to our proposed practice by a clear margin. This result indicates the necessity of designing highly semantics-correlated reward signals in our proposed method. In our proposed \ours~model, the tokens corresponding to task and coordinate prompts are relatively easy to be learned upon our observation, as they appear in the decoded sequence in a fixed order. Besides, the coordinate values are not directly determined by these tokens so that it's not suitable to share the same combinational semantics over all tokens.

\begin{table}[t]
  \centering
  \caption{Comparison results with traditional (non-UI customized) SOTA grounding approaches.}
  \resizebox{\textwidth}{!}{
    \begin{tabular}{lcccccccc}
    \toprule
    \specialrule{0em}{1.5pt}{1pt}   
    \multicolumn{1}{c}{\multirow{3}[0]{*}{\textbf{Models}}} & \multicolumn{4}{c}{\textbf{Mobile Data}} & \multicolumn{4}{c}{\textbf{Desktop Data}} \\
    \multicolumn{1}{c}{} & \multicolumn{2}{c}{\textbf{App Seen}} & \multicolumn{2}{c}{\textbf{App Unseen}} & \multicolumn{2}{c}{\textbf{Web Seen}} & \multicolumn{2}{c}{\textbf{Web Unseen}} \\
    \multicolumn{1}{c}{} & mIoU  & Acc (\%) & mIoU  & Acc (\%) & mIoU  & Acc (\%)  & mIoU  & Acc (\%) \\
    \hline
    \specialrule{0em}{1.5pt}{1pt}   
    GLIP (original) & 0.03 & 8.64 & 0.03 & 7.02 & 0.01 & 2.23 & 0.01 & 2.72 \\
    Grounding-DINO (original) & 0.07 & 10.31 & 0.05 & 8.97 & 0.03 & 4.25 & 0.03 & 3.87 \\
    GLIP (trained on UI data) & 0.18 & 20.36 & 0.12 & 14.91 & 0.07 & 9.54 & 0.06 & 8.75 \\
    Grounding-DINO (trained on UI data) & 0.27 & 28.29 & 0.23 & 23.83 & 0.21 & 20.06 & 0.19 & 18.62 \\
    RUIG (ours)  & 0.62 & 81.16 & 0.48 & 73.92 & 0.51 & 61.78 & 0.49 & 59.03 \\
    \bottomrule
    \end{tabular}
    }
  \label{tab:nongeneralist}%
\end{table}%

\subsection{Comparison with the State-of-the-Arts}

\begin{table}[!t]
  \centering
  \begin{minipage}{0.47\linewidth}
      \centering
      \caption{Investigation results on whether the policy gradients based loss should be adopted to all tokens. In \textit{RUIG (all tokens)}, we back-propagate the IoU-based rewards as supervisions for all tokens. In \textit{RUIG (proposed)}, we sorely back-propagate them for the tokens corresponding to coordinate values.}
      \resizebox{\textwidth}{!}{
        \begin{tabular}{lcccc}
        \toprule
        \multicolumn{1}{c}{\multirow{2}[0]{*}{\textbf{Models}}} & \multicolumn{2}{c}{\textbf{App Seen}} & \multicolumn{2}{c}{\textbf{App Unseen}} \\
              & mIoU  & Acc (\%) & mIoU  & Acc (\%) \\
        \hline
        \specialrule{0em}{1.5pt}{1pt}   
        Baseline & 0.52  & 72.23 & 0.42  & 65.03 \\
        RUIG (all tokens) & 0.54  & 76.65 & 0.43  & 69.12 \\
        RUIG (proposed) & 0.62  & 81.16 & 0.48  & 73.92 \\
        \bottomrule
        \end{tabular}%
        }
      \label{tab:maskedPG}%
  \end{minipage}
  \hfill
  \begin{minipage}{0.51\linewidth}
      \centering
      \caption{Comparison results (Acc, \%) with the state-of-the-art UI-tailored approaches on instruction grounding. Here, the Spotlight* \citep{li2022spotlight} is the one reproduced with the same training and testing configurations with ours.}
      \resizebox{\textwidth}{!}{
        \begin{tabular}{lcc}
        \toprule
        \multicolumn{1}{c}{\textbf{Models}} & \textbf{App Seen} & \textbf{App Unseen} \\
        \hline
        \specialrule{0em}{1.5pt}{1pt}   
        Seq2Seq \citep{shridhar2020alfred} & 40.40  & 31.30 \\
        MOCA \citep{singh2021factorizing} & 40.00    & 32.70 \\
        Seq2Act \citep{li2020mapping} & 64.40  & 62.20 \\
        Spotlight* \citep{li2022spotlight} & 76.83 & 68.76 \\
        RUIG (Ours) & 81.16 & 73.92 \\
        \bottomrule
        \end{tabular}%
      }
      \label{tab:sota}%
  \end{minipage}
\end{table}

\paragraph{Comparison with traditional grounding approaches.}

We experimentally compare our proposed \ours~model to non-UI customized approaches GLIP~\citep{li2022grounded}, Grounding-DINO~\citep{liu2023grounding} and their fine-tuned versions trained on our UI data. As shown in Table~\ref{tab:nongeneralist}, these traditional grounding approaches are significantly inferior to ours across different experimental settings. This can be attributed to two main reasons: 1) These approaches lack sufficient OCR capability. UI pages contain dense texts, necessitating the OCR capability for grounding UI elements upon task instructions. 2) They cannot fully comprehend natural language instructions. In specific, GLIP only support word prompts while Grounding DINO support sub-sentence prompts.

\paragraph{Comparison with UI-tailored grounding approaches.}

We compare our proposed \ours~model with the state-of-the-art UI-tailored approaches on the public benchmark proposed in \cite{burns2022dataset}. The results are in Table \ref{tab:sota}. Note that the works Seq2Seq \citep{shridhar2020alfred}, MOCA \citep{singh2021factorizing} and Seq2Act \citep{li2020mapping} all use the metadata of UIs, \ieno, view hierarchies. In Seq2Act \citep{li2020mapping}, a phrase extraction model is trained to explicitly parse each input instruction into its operation, object and additional arguments. Differently, our model allows to directly take natural instruction sentences as the inputs. The Spotlight* refers to the reproduced version of the model in \cite{li2022spotlight}, where we train Spotlight model using the same training configurations as we use to train our model. 
This model predicts YES or NO probability for each UI element and take the element with the largest probability for YES token as the grounding result. It thus requires the bounding boxes of all UI elements as the prior, where we use the bounding boxes provided by view hierarchies when re-train the model on this benchmark dataset.

As shown in Table \ref{tab:sota}, our proposed \ours~model achieves the best accuracy on both App Seen and App Unseen settings in comparisons with other methods. It is a pure-vision solution, obviating the need of using metadata or additional information (\egno, bounding boxes of UI elements). Thus, it exhibits impressive potentials of serving as a generic UI task automation API.

\subsection{Visualization Results}

\begin{figure}[t]
	\begin{center}
		\includegraphics[width=0.925\textwidth]{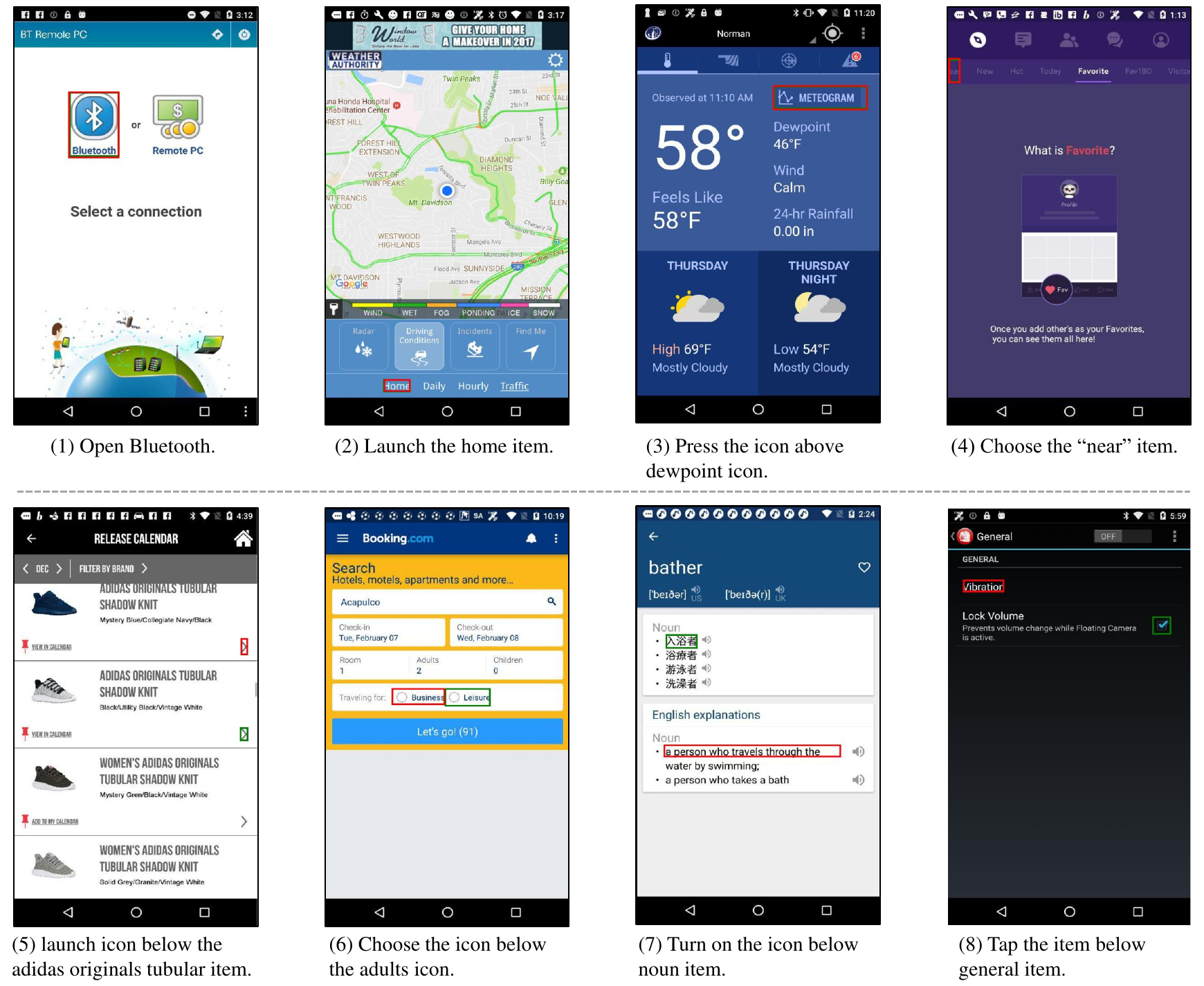} 
	\end{center}
	\caption{The visualization results of the grounded bounding boxes. The top row shows successful cases while the bottom row shows failure cases. Given instructions are under their corresponding screenshots. The model outputs are displayed in red, and the labels are shown in green. 
 }
    \vspace{-4mm}
	\label{fig:visualization}
\end{figure}

We visualize the predicted bounding boxes of our proposed \ours~model to show its capacity and analyze its failure cases in Figure \ref{fig:visualization}. Here, we present the results on mobile data for better visibility, considering the UI elements in desktop data are relative small. The successful cases shown in the top row of Figure \ref{fig:visualization} demonstrate our proposed \ours~model is competent for localizing the UI elements at different scales and performing grounding upon between-element relations. The case (4) exhibits that it can find partially occluded UI element in the background with a confused color. The failure cases actually seem reasonable in line with human understandings. The cases (5) (6) and (7) indicate the label ambiguity and the case (8) exposes the noisy labels in the currently used dataset.

\section{Conclusion}

In this paper, we construct a powerful UI instruction grounding model, named \ours. This model only requires natural instructions and screenshots as its inputs without the need of using metadata or additional information as previous works require. To achieve this, we cast instruction grounding as a promptable detection task, and adopt \textit{pixel-to-sequence} paradigm to localize the target UI element in linguistic form. This paradigm allows us to exploit the pre-trained knowledge from other image-to-text task. Moreover, we improve vanilla \textit{pixel-to-sequence} model by endowing it with the awareness of combinational semantics during its training, through our proposed policy gradients based optimization. Extensive experiments show our proposed method deliver significant performance improvements. As for the broad impact, from the perspective of model functionality, this work shows promises in building generic UI task automation APIs where LLMs serve as planners while domain-specific models/APIs function as executors. From the perspective of methodology, our proposed modification for \textit{pixel-to-sequence} paradigm is generally applicable for other tasks, and we hope it can inspire more excellent works in the future.

\bibliography{iclr2024_conference}

\begin{thebibliography}{50}
\providecommand{\natexlab}[1]{#1}
\providecommand{\url}[1]{\texttt{#1}}
\expandafter\ifx\csname urlstyle\endcsname\relax
  \providecommand{\doi}[1]{doi: #1}\else
  \providecommand{\doi}{doi: \begingroup \urlstyle{rm}\Url}\fi

\bibitem[Bai et~al.(2021)Bai, Zang, Xu, Sunkara, Rastogi, Chen, et~al.]{bai2021uibert}
Chongyang Bai, Xiaoxue Zang, Ying Xu, Srinivas Sunkara, Abhinav Rastogi, Jindong Chen, et~al.
\newblock Uibert: Learning generic multimodal representations for ui understanding.
\newblock \emph{arXiv preprint arXiv:2107.13731}, 2021.

\bibitem[Burns et~al.(2022)Burns, Arsan, Agrawal, Kumar, Saenko, and Plummer]{burns2022dataset}
Andrea Burns, Deniz Arsan, Sanjna Agrawal, Ranjitha Kumar, Kate Saenko, and Bryan~A Plummer.
\newblock A dataset for interactive vision-language navigation with unknown command feasibility.
\newblock In \emph{Computer Vision--ECCV 2022: 17th European Conference, Tel Aviv, Israel, October 23--27, 2022, Proceedings, Part VIII}, pp.\  312--328. Springer, 2022.

\bibitem[Chen et~al.(2022{\natexlab{a}})Chen, Saxena, Li, Fleet, and Hinton]{chen2021pix2seq}
Ting Chen, Saurabh Saxena, Lala Li, David~J Fleet, and Geoffrey Hinton.
\newblock Pix2seq: A language modeling framework for object detection.
\newblock \emph{ICLR}, 2022{\natexlab{a}}.

\bibitem[Chen et~al.(2022{\natexlab{b}})Chen, Saxena, Li, Lin, Fleet, and Hinton]{chen2022unified}
Ting Chen, Saurabh Saxena, Lala Li, Tsung-Yi Lin, David~J Fleet, and Geoffrey~E Hinton.
\newblock A unified sequence interface for vision tasks.
\newblock \emph{Advances in Neural Information Processing Systems}, 35:\penalty0 31333--31346, 2022{\natexlab{b}}.

\bibitem[Cho et~al.(2021)Cho, Lei, Tan, and Bansal]{cho2021unifying}
Jaemin Cho, Jie Lei, Hao Tan, and Mohit Bansal.
\newblock Unifying vision-and-language tasks via text generation.
\newblock In \emph{International Conference on Machine Learning}, pp.\  1931--1942. PMLR, 2021.

\bibitem[Chung et~al.(2022)Chung, Hou, Longpre, Zoph, Tay, Fedus, Li, Wang, Dehghani, Brahma, et~al.]{chung2022scaling}
Hyung~Won Chung, Le~Hou, Shayne Longpre, Barret Zoph, Yi~Tay, William Fedus, Eric Li, Xuezhi Wang, Mostafa Dehghani, Siddhartha Brahma, et~al.
\newblock Scaling instruction-finetuned language models.
\newblock \emph{arXiv preprint arXiv:2210.11416}, 2022.

\bibitem[Gravitas(2023)]{AutoGPT}
Significant Gravitas.
\newblock Auto-gpt, 2023.
\newblock \url{https://github.com/Significant-Gravitas/Auto-GPT#auto-gpt-an-autonomous-gpt-4-experiment}.

\bibitem[Gupta et~al.(2022)Gupta, Kamath, Kembhavi, and Hoiem]{gupta2022towards}
Tanmay Gupta, Amita Kamath, Aniruddha Kembhavi, and Derek Hoiem.
\newblock Towards general purpose vision systems: An end-to-end task-agnostic vision-language architecture.
\newblock In \emph{Proceedings of the IEEE/CVF Conference on Computer Vision and Pattern Recognition}, pp.\  16399--16409, 2022.

\bibitem[Gur et~al.(2018)Gur, Rueckert, Faust, and Hakkani-Tur]{gur2018learning}
Izzeddin Gur, Ulrich Rueckert, Aleksandra Faust, and Dilek Hakkani-Tur.
\newblock Learning to navigate the web.
\newblock \emph{arXiv preprint arXiv:1812.09195}, 2018.

\bibitem[He et~al.(2021)He, Sunkara, Zang, Xu, Liu, Wichers, Schubiner, Lee, and Chen]{he2021actionbert}
Zecheng He, Srinivas Sunkara, Xiaoxue Zang, Ying Xu, Lijuan Liu, Nevan Wichers, Gabriel Schubiner, Ruby Lee, and Jindong Chen.
\newblock Actionbert: Leveraging user actions for semantic understanding of user interfaces.
\newblock In \emph{Proceedings of the AAAI Conference on Artificial Intelligence}, volume~35, pp.\  5931--5938, 2021.

\bibitem[Humphreys et~al.(2022)Humphreys, Raposo, Pohlen, Thornton, Chhaparia, Muldal, Abramson, Georgiev, Goldin, Santoro, et~al.]{humphreys2022data}
Peter~C Humphreys, David Raposo, Toby Pohlen, Gregory Thornton, Rachita Chhaparia, Alistair Muldal, Josh Abramson, Petko Georgiev, Alex Goldin, Adam Santoro, et~al.
\newblock A data-driven approach for learning to control computers.
\newblock \emph{arXiv preprint arXiv:2202.08137}, 2022.

\bibitem[Iki \& Aizawa(2022)Iki and Aizawa]{iki2022berts}
Taichi Iki and Akiko Aizawa.
\newblock Do berts learn to use browser user interface? exploring multi-step tasks with unified vision-and-language berts.
\newblock \emph{arXiv preprint arXiv:2203.07828}, 2022.

\bibitem[Jang et~al.(2022)Jang, Kong, Jeon, Kim, and Kwak]{jang2022unifying}
Jiho Jang, Chaerin Kong, Donghyeon Jeon, Seonhoon Kim, and Nojun Kwak.
\newblock Unifying vision-language representation space with single-tower transformer.
\newblock \emph{arXiv preprint arXiv:2211.11153}, 2022.

\bibitem[Kim et~al.(2022)Kim, Hong, Yim, Nam, Park, Yim, Hwang, Yun, Han, and Park]{kim2022ocr}
Geewook Kim, Teakgyu Hong, Moonbin Yim, JeongYeon Nam, Jinyoung Park, Jinyeong Yim, Wonseok Hwang, Sangdoo Yun, Dongyoon Han, and Seunghyun Park.
\newblock Ocr-free document understanding transformer.
\newblock In \emph{Computer Vision--ECCV 2022: 17th European Conference, Tel Aviv, Israel, October 23--27, 2022, Proceedings, Part XXVIII}, pp.\  498--517. Springer, 2022.

\bibitem[Kim et~al.(2023)Kim, Baldi, and McAleer]{kim2023language}
Geunwoo Kim, Pierre Baldi, and Stephen McAleer.
\newblock Language models can solve computer tasks.
\newblock \emph{arXiv preprint arXiv:2303.17491}, 2023.

\bibitem[Le et~al.(2022)Le, Rathour, Yamazaki, Luu, and Savvides]{le2022deep}
Ngan Le, Vidhiwar~Singh Rathour, Kashu Yamazaki, Khoa Luu, and Marios Savvides.
\newblock Deep reinforcement learning in computer vision: a comprehensive survey.
\newblock \emph{Artificial Intelligence Review}, pp.\  1--87, 2022.

\bibitem[Lewis et~al.(2019)Lewis, Liu, Goyal, Ghazvininejad, Mohamed, Levy, Stoyanov, and Zettlemoyer]{lewis2019bart}
Mike Lewis, Yinhan Liu, Naman Goyal, Marjan Ghazvininejad, Abdelrahman Mohamed, Omer Levy, Ves Stoyanov, and Luke Zettlemoyer.
\newblock Bart: Denoising sequence-to-sequence pre-training for natural language generation, translation, and comprehension.
\newblock \emph{arXiv preprint arXiv:1910.13461}, 2019.

\bibitem[Li \& Li(2022)Li and Li]{li2022spotlight}
Gang Li and Yang Li.
\newblock Spotlight: Mobile ui understanding using vision-language models with a focus.
\newblock \emph{arXiv preprint arXiv:2209.14927}, 2022.

\bibitem[Li et~al.(2022)Li, Zhang, Zhang, Yang, Li, Zhong, Wang, Yuan, Zhang, Hwang, et~al.]{li2022grounded}
Liunian~Harold Li, Pengchuan Zhang, Haotian Zhang, Jianwei Yang, Chunyuan Li, Yiwu Zhong, Lijuan Wang, Lu~Yuan, Lei Zhang, Jenq-Neng Hwang, et~al.
\newblock Grounded language-image pre-training.
\newblock In \emph{Proceedings of the IEEE/CVF Conference on Computer Vision and Pattern Recognition}, pp.\  10965--10975, 2022.

\bibitem[Li et~al.(2020{\natexlab{a}})Li, He, Zhou, Zhang, and Baldridge]{li2020mapping}
Yang Li, Jiacong He, Xin Zhou, Yuan Zhang, and Jason Baldridge.
\newblock Mapping natural language instructions to mobile ui action sequences.
\newblock \emph{ACL}, 2020{\natexlab{a}}.

\bibitem[Li et~al.(2020{\natexlab{b}})Li, Li, He, Zheng, Li, and Guan]{li2020widget}
Yang Li, Gang Li, Luheng He, Jingjie Zheng, Hong Li, and Zhiwei Guan.
\newblock Widget captioning: Generating natural language description for mobile user interface elements.
\newblock \emph{arXiv preprint arXiv:2010.04295}, 2020{\natexlab{b}}.

\bibitem[Liang et~al.(2023)Liang, Wu, Song, Wu, Xia, Liu, Ou, Lu, Ji, Mao, et~al.]{liang2023taskmatrix}
Yaobo Liang, Chenfei Wu, Ting Song, Wenshan Wu, Yan Xia, Yu~Liu, Yang Ou, Shuai Lu, Lei Ji, Shaoguang Mao, et~al.
\newblock Taskmatrix. ai: Completing tasks by connecting foundation models with millions of apis.
\newblock \emph{arXiv preprint arXiv:2303.16434}, 2023.

\bibitem[Lin et~al.(2021)Lin, Zhang, Li, Zeng, and Chen]{lin2021selectaugment}
Shiqi Lin, Zhizheng Zhang, Xin Li, Wenjun Zeng, and Zhibo Chen.
\newblock Selectaugment: Hierarchical deterministic sample selection for data augmentation.
\newblock \emph{arXiv preprint arXiv:2112.02862}, 2021.

\bibitem[Liu et~al.(2018)Liu, Guu, Pasupat, Shi, and Liang]{liu2018reinforcement}
Evan~Zheran Liu, Kelvin Guu, Panupong Pasupat, Tianlin Shi, and Percy Liang.
\newblock Reinforcement learning on web interfaces using workflow-guided exploration.
\newblock \emph{arXiv preprint arXiv:1802.08802}, 2018.

\bibitem[Liu et~al.(2023)Liu, Zeng, Ren, Li, Zhang, Yang, Li, Yang, Su, Zhu, et~al.]{liu2023grounding}
Shilong Liu, Zhaoyang Zeng, Tianhe Ren, Feng Li, Hao Zhang, Jie Yang, Chunyuan Li, Jianwei Yang, Hang Su, Jun Zhu, et~al.
\newblock Grounding dino: Marrying dino with grounded pre-training for open-set object detection.
\newblock \emph{arXiv preprint arXiv:2303.05499}, 2023.

\bibitem[Liu et~al.(2021)Liu, Lin, Cao, Hu, Wei, Zhang, Lin, and Guo]{liu2021swin}
Ze~Liu, Yutong Lin, Yue Cao, Han Hu, Yixuan Wei, Zheng Zhang, Stephen Lin, and Baining Guo.
\newblock Swin transformer: Hierarchical vision transformer using shifted windows.
\newblock In \emph{Proceedings of the IEEE/CVF international conference on computer vision}, pp.\  10012--10022, 2021.

\bibitem[Mathe et~al.(2016)Mathe, Pirinen, and Sminchisescu]{mathe2016reinforcement}
Stefan Mathe, Aleksis Pirinen, and Cristian Sminchisescu.
\newblock Reinforcement learning for visual object detection.
\newblock In \emph{Proceedings of the IEEE conference on computer vision and pattern recognition}, pp.\  2894--2902, 2016.

\bibitem[OpenAI(2023{\natexlab{a}})]{ChatGPT}
OpenAI.
\newblock Introducing chatgpt, 2023{\natexlab{a}}.
\newblock \url{https://openai.com/blog/chatgpt}.

\bibitem[OpenAI(2023{\natexlab{b}})]{ChatGPTplugins}
OpenAI.
\newblock Chatgpt plugins, 2023{\natexlab{b}}.
\newblock \url{https://openai.com/blog/chatgpt-plugins}.

\bibitem[OpenAI(2023{\natexlab{c}})]{GPT4}
OpenAI.
\newblock Gpt-4, 2023{\natexlab{c}}.
\newblock \url{https://openai.com/research/gpt-4}.

\bibitem[Ouyang et~al.(2022)Ouyang, Wu, Jiang, Almeida, Wainwright, Mishkin, Zhang, Agarwal, Slama, Ray, et~al.]{ouyang2022training}
Long Ouyang, Jeffrey Wu, Xu~Jiang, Diogo Almeida, Carroll Wainwright, Pamela Mishkin, Chong Zhang, Sandhini Agarwal, Katarina Slama, Alex Ray, et~al.
\newblock Training language models to follow instructions with human feedback.
\newblock \emph{Advances in Neural Information Processing Systems}, 35:\penalty0 27730--27744, 2022.

\bibitem[Pinto et~al.(2023)Pinto, Kolesnikov, Shi, Beyer, and Zhai]{pinto2023tuning}
Andr{\'e}~Susano Pinto, Alexander Kolesnikov, Yuge Shi, Lucas Beyer, and Xiaohua Zhai.
\newblock Tuning computer vision models with task rewards.
\newblock \emph{arXiv preprint arXiv:2302.08242}, 2023.

\bibitem[Ramamurthy et~al.(2022)Ramamurthy, Ammanabrolu, Brantley, Hessel, Sifa, Bauckhage, Hajishirzi, and Choi]{ramamurthy2022reinforcement}
Rajkumar Ramamurthy, Prithviraj Ammanabrolu, Kiant{\'e} Brantley, Jack Hessel, Rafet Sifa, Christian Bauckhage, Hannaneh Hajishirzi, and Yejin Choi.
\newblock Is reinforcement learning (not) for natural language processing?: Benchmarks, baselines, and building blocks for natural language policy optimization.
\newblock \emph{arXiv preprint arXiv:2210.01241}, 2022.

\bibitem[reworkd.ai(2023)]{AgentGPT}
reworkd.ai.
\newblock Agentgpt, 2023.
\newblock \url{https://github.com/reworkd/AgentGPT}.

\bibitem[Shen et~al.(2023)Shen, Song, Tan, Li, Lu, and Zhuang]{shen2023hugginggpt}
Yongliang Shen, Kaitao Song, Xu~Tan, Dongsheng Li, Weiming Lu, and Yueting Zhuang.
\newblock Hugginggpt: Solving ai tasks with chatgpt and its friends in huggingface.
\newblock \emph{arXiv preprint arXiv:2303.17580}, 2023.

\bibitem[Shi et~al.(2017)Shi, Karpathy, Fan, Hernandez, and Liang]{shi2017world}
Tianlin Shi, Andrej Karpathy, Linxi Fan, Jonathan Hernandez, and Percy Liang.
\newblock World of bits: An open-domain platform for web-based agents.
\newblock In \emph{ICML}, pp.\  3135--3144, 2017.

\bibitem[Shridhar et~al.(2020)Shridhar, Thomason, Gordon, Bisk, Han, Mottaghi, Zettlemoyer, and Fox]{shridhar2020alfred}
Mohit Shridhar, Jesse Thomason, Daniel Gordon, Yonatan Bisk, Winson Han, Roozbeh Mottaghi, Luke Zettlemoyer, and Dieter Fox.
\newblock Alfred: A benchmark for interpreting grounded instructions for everyday tasks.
\newblock In \emph{Proceedings of the IEEE/CVF conference on computer vision and pattern recognition}, pp.\  10740--10749, 2020.

\bibitem[Singh et~al.(2021)Singh, Bhambri, Kim, Mottaghi, and Choi]{singh2021factorizing}
Kunal~Pratap Singh, Suvaansh Bhambri, Byeonghwi Kim, Roozbeh Mottaghi, and Jonghyun Choi.
\newblock Factorizing perception and policy for interactive instruction following.
\newblock In \emph{Proceedings of the IEEE/CVF International Conference on Computer Vision}, pp.\  1888--1897, 2021.

\bibitem[Sutton \& Barto(2018)Sutton and Barto]{sutton2018reinforcement}
Richard~S Sutton and Andrew~G Barto.
\newblock \emph{Reinforcement learning: An introduction}.
\newblock MIT press, 2018.

\bibitem[Touvron et~al.(2023)Touvron, Lavril, Izacard, Martinet, Lachaux, Lacroix, Rozi{\`e}re, Goyal, Hambro, Azhar, et~al.]{touvron2023llama}
Hugo Touvron, Thibaut Lavril, Gautier Izacard, Xavier Martinet, Marie-Anne Lachaux, Timoth{\'e}e Lacroix, Baptiste Rozi{\`e}re, Naman Goyal, Eric Hambro, Faisal Azhar, et~al.
\newblock Llama: Open and efficient foundation language models.
\newblock \emph{arXiv preprint arXiv:2302.13971}, 2023.

\bibitem[Uc-Cetina et~al.(2023)Uc-Cetina, Navarro-Guerrero, Martin-Gonzalez, Weber, and Wermter]{uc2023survey}
Victor Uc-Cetina, Nicolas Navarro-Guerrero, Anabel Martin-Gonzalez, Cornelius Weber, and Stefan Wermter.
\newblock Survey on reinforcement learning for language processing.
\newblock \emph{Artificial Intelligence Review}, 56\penalty0 (2):\penalty0 1543--1575, 2023.

\bibitem[Vaswani et~al.(2017)Vaswani, Shazeer, Parmar, Uszkoreit, Jones, Gomez, Kaiser, and Polosukhin]{vaswani2017attention}
Ashish Vaswani, Noam Shazeer, Niki Parmar, Jakob Uszkoreit, Llion Jones, Aidan~N Gomez, {\L}ukasz Kaiser, and Illia Polosukhin.
\newblock Attention is all you need.
\newblock \emph{Advances in neural information processing systems}, 30, 2017.

\bibitem[Vemprala et~al.(2023)Vemprala, Bonatti, Bucker, and Kapoor]{vemprala2023chatgpt}
Sai Vemprala, Rogerio Bonatti, Arthur Bucker, and Ashish Kapoor.
\newblock Chatgpt for robotics: Design principles and model abilities.
\newblock Technical Report MSR-TR-2023-8, Microsoft, February 2023.
\newblock URL \url{https://www.microsoft.com/en-us/research/publication/chatgpt-for-robotics-design-principles-and-model-abilities/}.

\bibitem[Williams \& Zipser(1989)Williams and Zipser]{williams1989learning}
Ronald~J Williams and David Zipser.
\newblock A learning algorithm for continually running fully recurrent neural networks.
\newblock \emph{Neural computation}, 1\penalty0 (2):\penalty0 270--280, 1989.

\bibitem[Wu et~al.(2023)Wu, Yin, Qi, Wang, Tang, and Duan]{wu2023visual}
Chenfei Wu, Shengming Yin, Weizhen Qi, Xiaodong Wang, Zecheng Tang, and Nan Duan.
\newblock Visual chatgpt: Talking, drawing and editing with visual foundation models.
\newblock \emph{arXiv preprint arXiv:2303.04671}, 2023.

\bibitem[Xu et~al.(2020)Xu, Li, Cui, Huang, Wei, and Zhou]{xu2020layoutlm}
Yiheng Xu, Minghao Li, Lei Cui, Shaohan Huang, Furu Wei, and Ming Zhou.
\newblock Layoutlm: Pre-training of text and layout for document image understanding.
\newblock In \emph{Proceedings of the 26th ACM SIGKDD International Conference on Knowledge Discovery \& Data Mining}, pp.\  1192--1200, 2020.

\bibitem[Yang et~al.(2022)Yang, Gan, Wang, Hu, Ahmed, Liu, Lu, and Wang]{yang2022unitab}
Zhengyuan Yang, Zhe Gan, Jianfeng Wang, Xiaowei Hu, Faisal Ahmed, Zicheng Liu, Yumao Lu, and Lijuan Wang.
\newblock Unitab: Unifying text and box outputs for grounded vision-language modeling.
\newblock In \emph{Computer Vision--ECCV 2022: 17th European Conference, Tel Aviv, Israel, October 23--27, 2022, Proceedings, Part XXXVI}, pp.\  521--539. Springer, 2022.

\bibitem[Yang et~al.(2023)Yang, Li, Wang, Lin, Azarnasab, Ahmed, Liu, Liu, Zeng, and Wang]{yang2023mm}
Zhengyuan Yang, Linjie Li, Jianfeng Wang, Kevin Lin, Ehsan Azarnasab, Faisal Ahmed, Zicheng Liu, Ce~Liu, Michael Zeng, and Lijuan Wang.
\newblock Mm-react: Prompting chatgpt for multimodal reasoning and action.
\newblock \emph{arXiv preprint arXiv:2303.11381}, 2023.

\bibitem[Zhang et~al.(2022)Zhang, Roller, Goyal, Artetxe, Chen, Chen, Dewan, Diab, Li, Lin, et~al.]{zhang2022opt}
Susan Zhang, Stephen Roller, Naman Goyal, Mikel Artetxe, Moya Chen, Shuohui Chen, Christopher Dewan, Mona Diab, Xian Li, Xi~Victoria Lin, et~al.
\newblock Opt: Open pre-trained transformer language models.
\newblock \emph{arXiv preprint arXiv:2205.01068}, 2022.

\bibitem[Zhou et~al.(2019)Zhou, Fang, Song, Guan, Yin, Dai, and Yang]{zhou2019iou}
Dingfu Zhou, Jin Fang, Xibin Song, Chenye Guan, Junbo Yin, Yuchao Dai, and Ruigang Yang.
\newblock Iou loss for 2d/3d object detection.
\newblock In \emph{2019 International Conference on 3D Vision (3DV)}, pp.\  85--94. IEEE, 2019.

\end{thebibliography}
\bibliographystyle{iclr2024_conference}

\clearpage

\appendix

\section{More Implementation Details}

We describe the primary implementation details in Sec.4.1 of the main body, and further provide additional details here. We follow the original Transformer \citep{vaswani2017attention} to adopt a teacher-forcing scheme \citep{williams1989learning} for model training, in which the ground truths are given as model inputs corresponding to the previous steps during the training of autoregressive decoding. For different training samples, the decoded sequence is generally organized as ``\textit{<instruction> \{content\} </instruction> <predict\_bbox> <x\_min> \{$x_{min}$\} </x\_min> <y\_min> \{$y_{min}$\} </y\_min> <x\_max> \{$x_{max}$\} </x\_max> <y\_max> \{$y_{max}$\} </y\_max></predict\_bbox> <eos>}''. Here, the tokens corresponding to ``\textit{<instruction> \{content\} </instruction>''} are masked out for discarding the supervisions on them, as they are user inputs. For all models, we adopt a half-percision training, and apply a gradient clipping technique whose maximum gradient norm is 1.0. The maximum length of the decoded sequence is set to 128.

\section{More Experiment Results}

\paragraph{Can the benefits of our proposed method be maintained when the model size is scaled up?}

Our proposed optimization method enables task-aligned supervision when decoding vision-related signals, which is theoretically applicable to models of different sizes. We believe that a more rational optimization approach can enhance the performance of models with varying sizes, and further conduct experiments to validate this. As presented in the table below, the benefits brought by the proposed optimization method remain significant when scaling up the size of the language decoder.

\begin{table}[h]
  \centering
  \caption{The performance of our proposed \ours~model when the model size is scaled up.}
  \resizebox{\textwidth}{!}{
    \begin{tabular}{lcccccccc}
    \toprule
    \specialrule{0em}{1.5pt}{1pt}   
    \multicolumn{1}{c}{\multirow{3}[0]{*}{\textbf{Models}}} & \multicolumn{4}{c}{\textbf{Mobile Data}} & \multicolumn{4}{c}{\textbf{Desktop Data}} \\
    \multicolumn{1}{c}{} & \multicolumn{2}{c}{\textbf{App Seen}} & \multicolumn{2}{c}{\textbf{App Unseen}} & \multicolumn{2}{c}{\textbf{Web Seen}} & \multicolumn{2}{c}{\textbf{Web Unseen}} \\
    \multicolumn{1}{c}{} & mIoU  & Acc (\%) & mIoU  & Acc (\%) & mIoU  & Acc (\%)  & mIoU  & Acc (\%) \\
    \hline
    \specialrule{0em}{1.5pt}{1pt}   
    Baseline (4 decoder layers) & 0.52 & 72.23 & 0.42 & 65.03 & 0.45 & 48.69 & 0.41 & 46.46 \\
    Our RUIG (4 decoder layers) & 0.62 & 81.16 & 0.48 & 73.92 & 0.51 & 61.78 & 0.49 & 59.03 \\
    Baseline (12 decoder layers) & 0.54 & 76.84 & 0.44 & 68.19 & 0.47 & 54.92 & 0.42 & 51.66 \\
    Our RUIG (12 decoder layers) & 0.65 & 83.99 & 0.51 & 77.30 & 0.53 & 65.37 & 0.52 & 65.17 \\
    \bottomrule
    \end{tabular}
    }
  \label{tab:nongeneralist}%
\end{table}%

\paragraph{Hyper-parameter choices when adopting policy gradients.}

\begin{wrapfigure}{r}{0.5\textwidth}
    \vspace{-10mm}
	\begin{center}
		\includegraphics[width=0.5\textwidth]{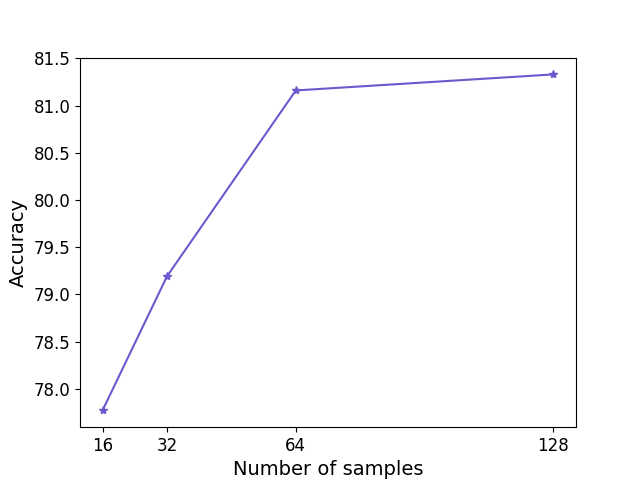} 
	\end{center}
    \vspace{-2mm}
	\caption{The experiment results (Acc, \%) for our proposed \ours~model with different Monte Carlo sampling times per each expectation estimation on mobile data (App Seen).}
    \vspace{-4mm}
	\label{fig:num_samples}
\end{wrapfigure}

We follow the common practice in RL field to perform estimation for each expectation item in Eq. \ref{eq:3} via Monte Carlo sampling with respect to the output logits for each token. In this part, we investigate the hyper-parameter choice for the Monte Carlo sampling times. The result on mobile data under the App Seen setting is shown in Figure \ref{fig:num_samples}. Similar experiment tendencies are observed on desktop data and other settings, thus ommited here for brevity. In theory, the more we sample, the more accurate the estimation of mathematical expectation is. In practice, we choose 64 as the default value in our experiment considering the training efficiency. With this hyper-parameter setting, our proposed \ours~model's training time per epoch is increased by 38\% on average, related to the baseline model. It remains almost the same convergence speed with the baseline model, indicating the using of combinational semantics facilitate the model convergence.

\section{More Visualization Results on Desktop Data}

\begin{figure}[th]
	\begin{center}
            \includegraphics[width=\textwidth]{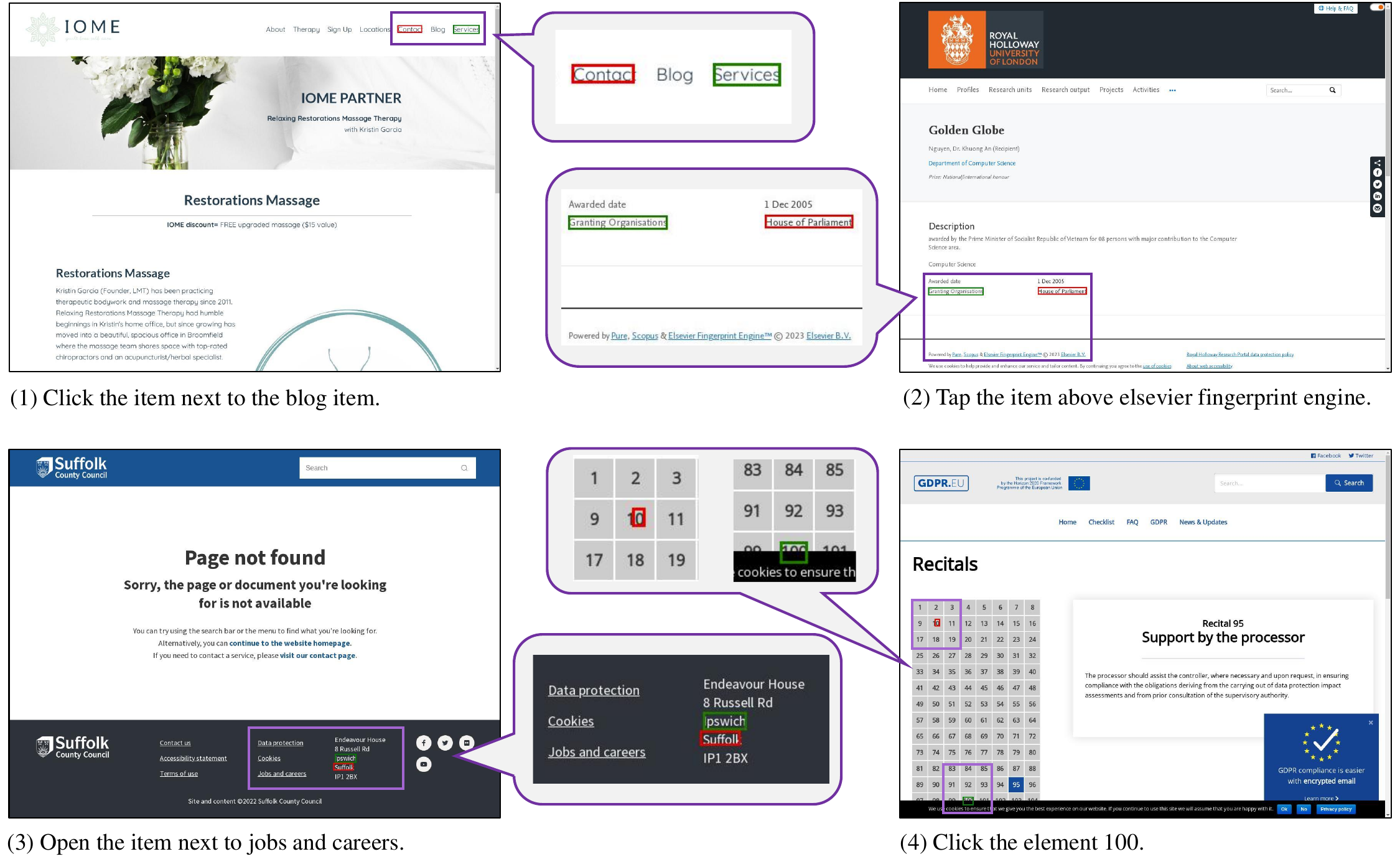}
	\end{center}
	\caption{The visualization results of the predicted bounding boxes on desktop data for failure cases. For each case, the instruction is provided below its corresponding screenshot. The predicted boxes are depicted in red while the ground truth boxes are depicted in green.}
	\label{fig:vis_failure}
\end{figure}

\begin{figure}[t]
	\begin{center}
            \includegraphics[width=\textwidth]{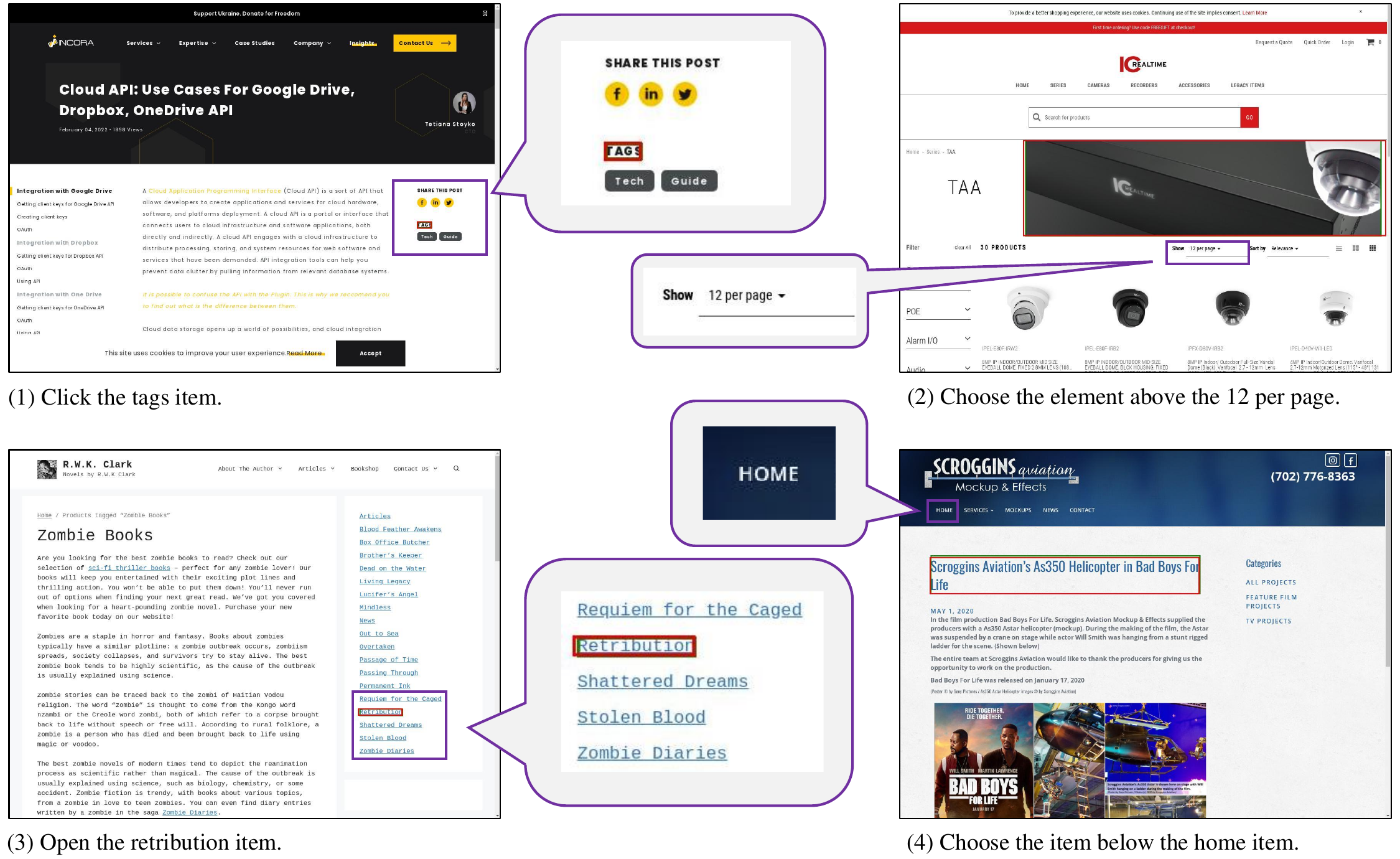} 
	\end{center}
    \vspace{-2mm}
	\caption{The visualization results of the predicted bounding boxes on desktop data for successful cases. For each case, the instruction is provided below its corresponding screenshot. The predicted boxes are depicted in red while the ground truth boxes are depicted in green.}
    \vspace{-5mm}
	\label{fig:vis_success}
\end{figure}

In the main text of our paper, we present the visualization results on mobile data and analyze them. In this section, we further provide the visualization results using desktop data. Successful cases are illustrated in Figure \ref{fig:vis_success}, and failure cases are illustrated in Figure \ref{fig:vis_failure}.

When comparing the desktop screenshots visualized here with those in the main paper, we observe that UI instruction grounding on desktop data appears to be more challenging than on mobile data, as the UI elements in desktop screenshots are more densely packed and exhibit greater scale diversity. The visualization results in Figure \ref{fig:vis_success} demonstrate that our proposed \ours~model is also capable of locating the target elements of various scales on desktop data, based on their contents or the relative positional relationship between the target elements and other elements. This implies the potential of our proposed \ours~model in serving as a generic task automation executor across different devices.

We further analyze the failure cases on desktop data. As illustrated in Figure \ref{fig:vis_failure}, our proposed \ours~model cannot predict aligned outputs with the ground truth results when there are ambiguous instructions or occluded target UI element. In specific, for the cases (1) (2) and (3) in Figure \ref{fig:vis_failure}, the model outputs are actually reasonable as well, considering that the given instructions are ambiguous. For the case (4) in Figure \ref{fig:vis_failure}, the target UI element is partially occluded by a pop-up window. In this case, our model finds the element that is the most similar to the target one as its prediction result.

\section{Examples of Unavailable Metadata}

\begin{figure}[t]
	\begin{center}
		\includegraphics[width=\textwidth]{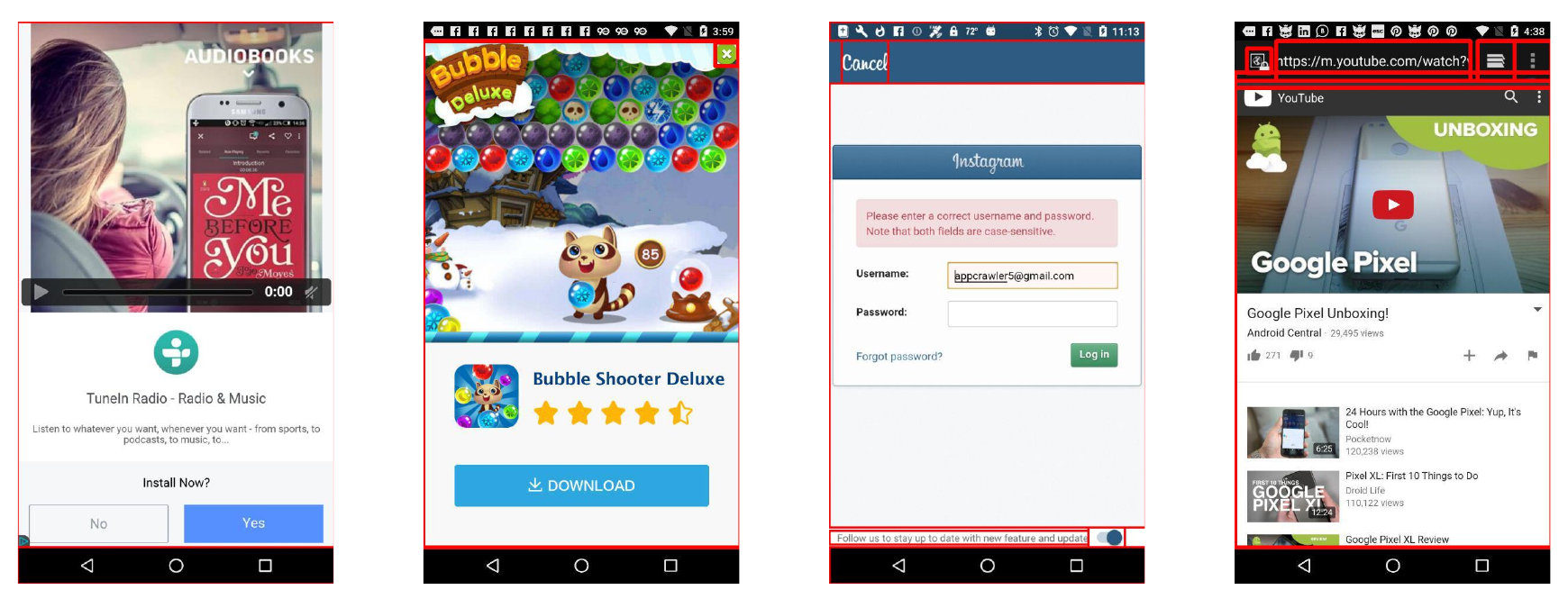} 
	\end{center}
	\caption{Examples of unavailable metadata. All elements available in the metadata are visualized in red bounding boxes. We can easily observe that the bounding box information of a considerable number of UI elements are not available in the corresponding metadata. 
 }
	\label{fig:visual_unavailable}
\end{figure}

We visualize examples of unavailable metadata in Figure \ref{fig:visual_unavailable}. We can easily observe that not all metadata for UI elements is readily available. To name a few, the ``Yes'' or ``No'' buttons in the first screenshot, the metadata of the ``DOWNLOAD'' button in the second screenshot, the ``Log in'' button in the third screenshot and the forward button in the fourth screenshot is all missed.

\section{Examples of Low-quality Metadata}

\begin{figure}[t]
	\begin{center}
		\includegraphics[width=\textwidth]{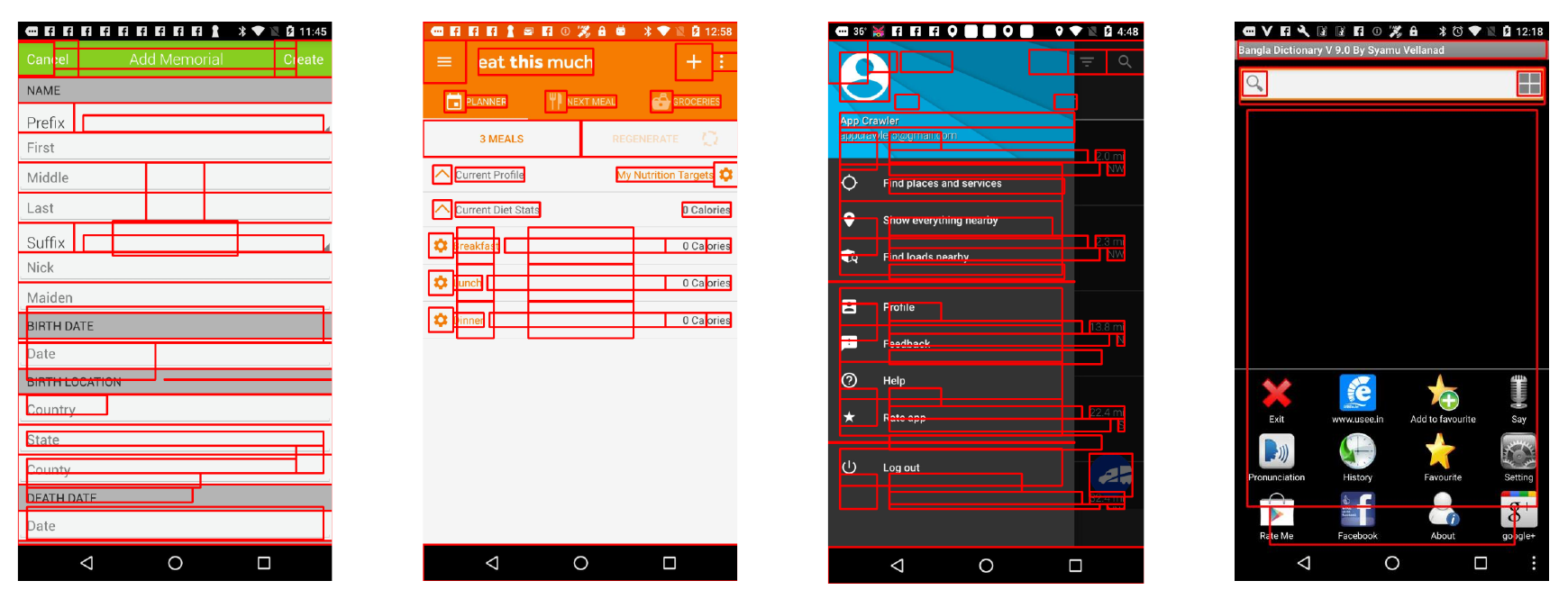} 
	\end{center}
	\caption{Examples of low-quality metadata. All elements available in the metadata are visualized in red bounding boxes. It can be easily observed that not all bounding boxes correspond to UI elements reasonably in the sense that some information in the metadata is noisy.
 }
	\label{fig:visual_lowquality}
\end{figure}

We visualize examples of low-quality metadata in Figure \ref{fig:visual_lowquality}. We can easily find that some bounding boxes in the metadata are chaotic. There are no UI elements corresponding to these disordered bounding boxes reasonably.

Note that the unavailable and low-quality metadata are both told at the UI element level, rather than at the screenshot level. We will clarify this in our revision.

\end{document}